\documentclass[runningheads]{llncs}

\usepackage{eccv}

\usepackage{eccvabbrv}

\usepackage{graphicx}
\usepackage{booktabs}

\usepackage[accsupp]{axessibility}  

\usepackage{hyperref}

\usepackage{orcidlink}
\usepackage[table]{xcolor}
\usepackage{lineno}
\usepackage[utf8]{inputenc} 
\usepackage[T1]{fontenc}    
\usepackage{url}            
\usepackage{booktabs}       
\usepackage{amsfonts}       
\usepackage{nicefrac}       
\usepackage{microtype}      
\usepackage{xcolor}         
\usepackage{enumitem}
\usepackage{listings}
\usepackage{amsmath, amssymb}
\usepackage{cuted}   
\usepackage{comment} 
\usepackage{textcomp}
\usepackage{tabularx}
\usepackage{listings}
\usepackage{tcolorbox}
\tcbuselibrary{skins,breakable}

\usepackage{graphicx}
\usepackage{amsmath}
\usepackage{amssymb}
\usepackage{bm}
\usepackage{multirow}
\usepackage{graphicx}
\usepackage{array}
\usepackage{booktabs}
\usepackage{makecell}   
\usepackage{caption}    
\usepackage[table]{xcolor}

\usepackage{subcaption} 
\usepackage{tablefootnote}
\usepackage{pifont} 

\begin{document}

\title{Teaching Vision-Language-Action Models What to See and Where to Look} 

\titlerunning{DriveTeach-VLA ECCV paper}

\author{
Yuguang Yang$^{*}$\inst{1,3}
\and
Canyu Chen$^{*}$\inst{2,3}
\and
Zhewen Tan$^{*}$\inst{6}
\and
Yizhi Wang\inst{7}
\and
Zichao Feng\inst{8}
\and
Chunyang Liu\inst{4}
\and
Kehua Sheng\inst{4}
\and
Juan Zhang\inst{8}
\and
Linlin Yang\inst{5}
\and
Baochang Zhang$^{\ddagger}$\inst{8}
\and
Yan Wang$^{\dagger}$\inst{3}
\and
Bo Zhang$^{\dagger}$\inst{4}
\and
Xianbin Cao$^{\dagger}$\inst{1}
}
\authorrunning{Y.~Yang et al.}

\institute{
\begin{tabular}[t]{l}
School of Electronic Information Engineering, Beihang University
\end{tabular}
\and
\begin{tabular}[t]{l}
National College for Excellent Engineers, Beihang University
\end{tabular}
\and
\begin{tabular}[t]{l}
Institute for AI Industry Research (AIR), Tsinghua University
\end{tabular}
\and
\begin{tabular}[t]{l}
DiDi
\end{tabular}
\and
\begin{tabular}[t]{l}
State Key Laboratory of Media Convergence and Communication,\\
Communication University of China
\end{tabular}
\and
\begin{tabular}[t]{l}
School of Computer Science and Engineering, Beihang University
\end{tabular}
\and
\begin{tabular}[t]{l}
School of Cyber Science and Technology, Beihang University
\end{tabular}
\and
\begin{tabular}[t]{l}
School of Artificial Intelligence, Beihang University
\end{tabular}
}

\maketitle
\begingroup
\renewcommand{\thefootnote}{}
\footnotetext{

$^{\ddagger}$ Project Lead.

$^{*}$ Equal contribution:
\url{{guangbuaa,chencanyu,tanzhewen}@buaa.edu.cn}

$^{\dagger}$ Corresponding authors:
\url{xbcao@buaa.edu.cn},
\url{wangyan@air.tsinghua.edu.cn},
\url{zhangbo@didiglobal.com}

}

\endgroup

\begin{abstract}
Vision-Language-Action (VLA) models have emerged as a promising paradigm for end-to-end autonomous driving. However, existing VLAs' traffic relies heavily on text-centric visual question answering and chain-of-thought reasoning data, which emphasizes linguistic reasoning rather than action-grounded planning. As a result, the learned representations capture semantic knowledge but lack spatial dependencies crucial for reliable trajectory prediction. We propose DriveTeach-VLA, a framework that explicitly teaches VLAs what to see and where to look. Driving-aware Vision Distillation (DVD) injects driving-specific perceptual priors into the vision encoder, while 2D Trajectory-Guided Prompts (2D-TGP) provide spatial conditioning aligned with feasible driving trajectories. Together they form a vision-guided learning pipeline: what to see (DVD pretraining) → where to look (TGP-guided SFT) → how to act (TGP-guided GRPO). DriveTeach-VLA achieves the state-of-the-art performance on NAVSIM and nuScenes. Our code is available at: \url{https://github.com/ShivaTeam/DriveTeach-VLA}.

\end{abstract}

\section{Introduction}
\label{sec:intro}
Autonomous Driving (AD) has long been a challenging application~\cite{broekman2025toward}. The previous dominant paradigm relied on a modular design, 
decomposing the driving process into 
{perception}~\cite{li2024bevformer, wang2022detr3d, liang2022bevfusion}, 
{prediction}~\cite{zhou2023qcnext, shi2024mtr++, huang2023gameformer}, and 
{planning}~\cite{huang2024gen, huang2023differentiable, liu2025hybrid} modules. 
While this structure offers interpretability and controllability, 
it suffers from error propagation and limited cross-task reasoning~\cite{zhou2025autovla}. Recent advances in Multi-modal Large Language Models (MLLMs)~\cite{InternVL, Qwen2.5, LLAVA} have demonstrated strong generalization and multi-task capabilities, inspiring a shift toward unified models that map raw sensor inputs to feasible trajectory planning results directly.

\begin{figure*}[t]
    \centering
    \includegraphics[width=1\linewidth]{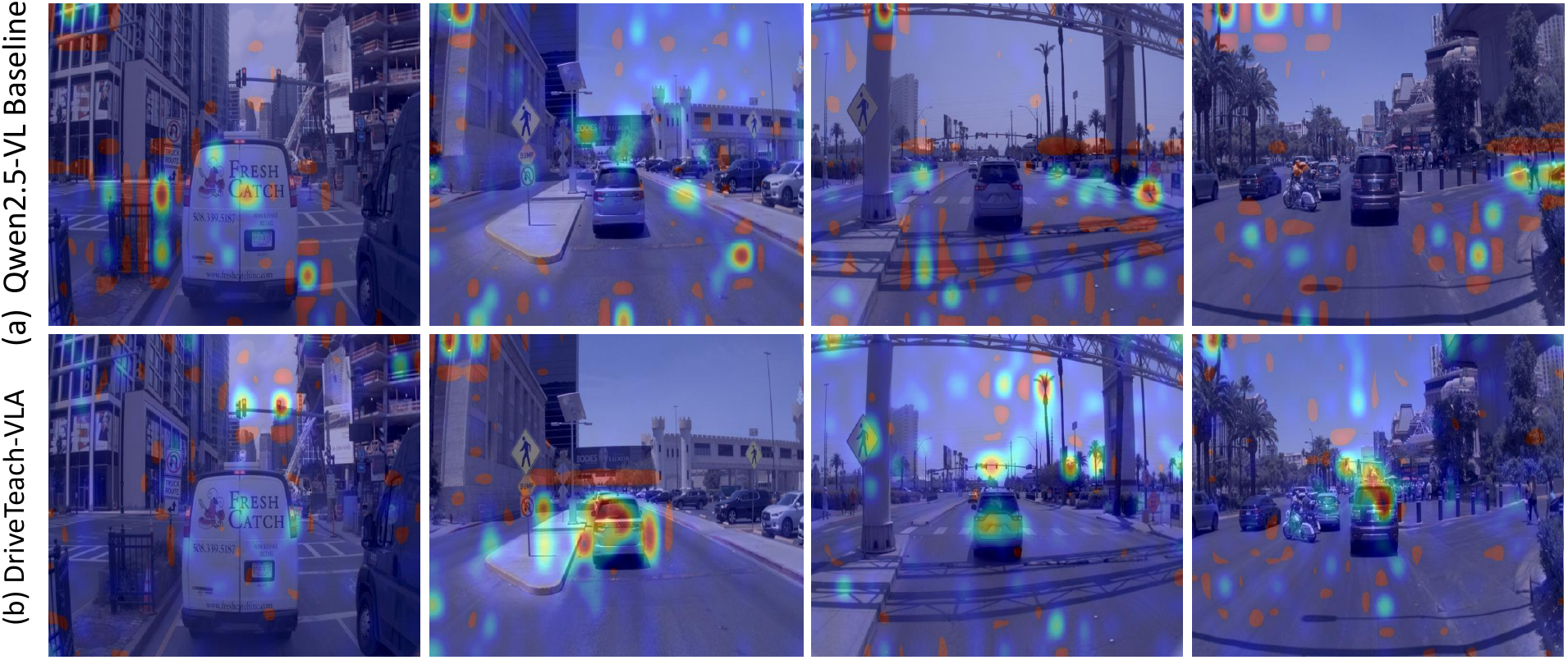}
    \caption{\textbf{Visualization of attention maps during autoregressive decoding on Qwen2.5-VL SFT baseline and DriveTeach-VLA.}
}
\label{fig:vit_attention_decode}
\end{figure*}

In this trend, {Vision–Language–Action (VLA)} models have emerged as a powerful framework, 
integrating perception, prediction, and planning in a unified manner~\cite{zhou2025opendrivevla, fu2025orion, li2025recogdrive, li2025drivevlaw0, liu2025occvla, zhou2025autovla}. 
These models typically take RGB camera images as visual inputs and predict the vehicle’s future trajectory over several seconds at 0.5\,s intervals. 
In terms of model design, the paradigm formulates trajectory prediction as a decoding task, where the MLLM-based planner either  auto-regressing text-based waypoints~\cite{chi2025impromptu, zhou2025autovla, hwang2024emma, xing2025openemma, luo2025adathinkdrive, rowe2025poutine}, or 
generating trajectories via an auxiliary planner$, e.g., $ a diffusion model, ~\cite{li2025recogdrive, fu2025orion, renz2025simlingo, yang2025drivemoe}.  In terms of training, recent advanced AD-VLA models~\cite{sima2024drivelm, renz2024carllava, renz2025simlingo, li2025recogdrive, luo2025adathinkdrive, zhou2025opendrivevla} typically adopt a three-stage training pipeline:  (\textit{i}) {Pretraining} on Visual Question Answer (VQA) data~\cite{sima2024drivelm, chi2025impromptu} to inject driving priors; (\textit{ii}) {Imitation Learning} (IL) with Supervised Fine Tuning (SFT) on Chain-of-Thought (CoT) reasoning data and expert trajectories. These two stages are common across nearly all multi-stage VLA frameworks.  (\textit{iii}) More recently, the {Reinforcement Learning} (RL) stage with GRPO has been introduced by currently advanced works~\cite{jiao2025evadrive, li2025recogdrive, zhou2025autovla, luo2025adathinkdrive} to further align AD-VLA's predicted trajectories with human driving preferences.

Although AD-VLAs can explain their intents, meta-behaviors, \textit{e.g.}, turning left/right, and even their trajectory planning decisions in a linguistically interpretable manner, a fundamental gap between perception and planning exists: Although VQA or CoT reasoning data introduces general traffic priors, textual contexts alone can not truly equip VLAs with the spatial and behavioral knowledge required for autonomous driving. This is due to their supervision being inherently \textit{text-centric}, focusing on semantic question answering rather than planning-related perception. Consequently, the learned representations emphasize linguistic semantics instead of motion and spatial dependencies essential for trajectory planning. Ultimately, the model learns to name  \textit{what is seen} rather than understand \textit{what should see} and \textit{where should look} for reliable trajectory generation. As shown in Figure~\ref{fig:vit_attention_decode}(a), attention learned from purely textual traffic-knowledge SFT is scattered across the scene with no clear spatial grounding, making it difficult to interpret. Quantitative analysis is further provided in Analysis~1 of Section~\ref{sec:ablation}.

\begin{figure*}[t]
    \centering
    \includegraphics[width=1\linewidth]{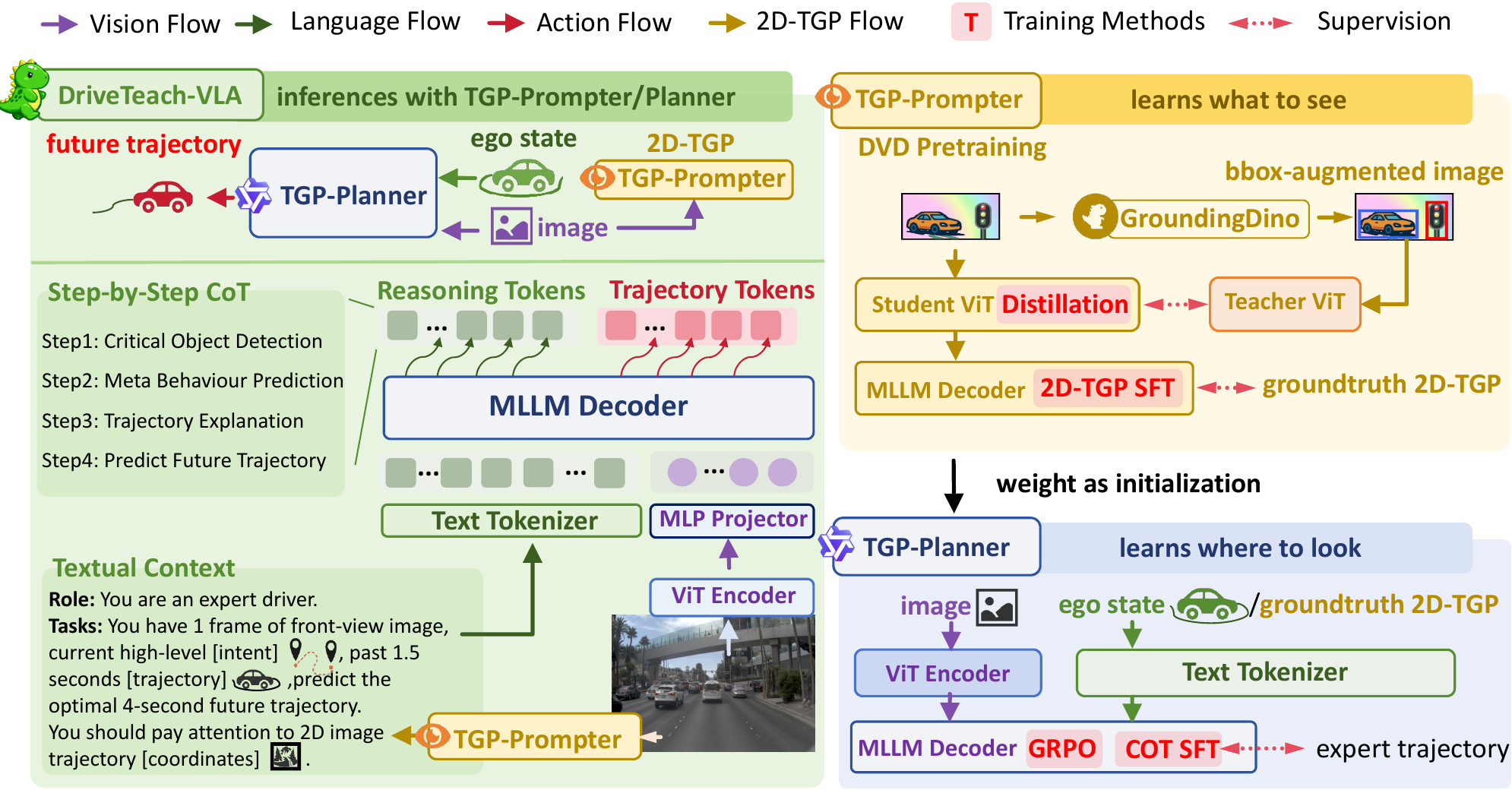}
    \caption{\textbf{DriveTeach-VLA Architecture.}
\textcolor{green!60!black}{\textbf{Left}}: DriveTeach-VLA operates in a dual-model manner consisting of a TGP-Prompter and a TGP-Planner. The TGP-Prompter first predicts the 2D-TGP, which is then used to condition the TGP-Planner for trajectory generation. 
\textcolor{orange!80!black}{\textbf{Right-Up}}: The TGP-Prompter is trained via DVD, supervised by critical-object bounding-box–augmented images' features and ground-truth 2D-TGP.  \textcolor{blue!70!black}{\textbf{Right-Down}}: The TGP-Planner is supervised by expert trajectories during IL with SFT and further optimized by RL with GRPO. During both IL and RL, ground-truth 2D-TGP is provided as contextual guidance (teacher forcing) to provide the precise planning-aware spatial guidance for trajectory learning.}
    \label{fig:pipeline}
\end{figure*}

In this paper, we propose {DriveTeach-VLA}, a framework that operates in a dual-model manner consisting of a TGP-Prompter and a TGP-Planner, both built upon Qwen2.5-VL-3B. The overall pipeline of DriveTeach-VLA is shown in Figure~\ref{fig:pipeline}. The TGP-Prompter undergoes vision-enhanced pretraining and is responsible for generating feasible driving region guidance, while TGP-Planner is conditioned on the guidance to predict future trajectory. 
To achieve \textbf{``teach what to see''}, \textit{i.e.}, which objects are relevant for driving, the TGP-Prompter is trained with Driving-aware Vision Distillation (DVD). Specifically, DVD first employs Grounding DINO~\cite{liu2024grounding} to detect traffic-critical objects. The detected bounding boxes are then overlaid on the images as visual prompts, forming so-called \textit{bbox-augmented images}. Then we apply self-distillation~\cite{caron2021emerging} on the ViT encoder of the TGP-Prompter: a teacher ViT processes bbox-augmented images while a student ViT processes the corresponding raw images. Their representations are aligned in the latent space, enabling the student encoder to internalize traffic-relevant visual cues without relying on textual supervision. To achieve \textbf{“teach where to look”}, \textit{i.e.}, which spatial region is most relevant for planning in the current scene, we further introduce a 2D image Trajectory-Guided Prompt (2D-TGP), which projects trajectories from world (BEV) coordinates onto the image plane with a pinhole camera model~\cite{heyden2005multiple}. These 2D-TGP trajectories can provide spatial guidance about feasible driving regions for the AD-VLA. Thus, the TGP-Prompter's decoder output is supervised with ground-truth 2D-TGP trajectory derived from expert trajectories via SFT. 

The TGP-Planner is initialized from the TGP-Prompter to obtain the traffic prior and trained to predict trajectories conditioned on the 2D-TGP. Following~\cite{chi2025impromptu, zhou2025autovla, rowe2025poutine}, we annotate step-by-step CoT reasoning data with Qwen2.5-VL-72B. The TGP-Planner learns to generate intermediate CoT reasoning steps and produce the future trajectory under the guidance of 2D-TGP. Training is performed with SFT followed by GRPO. To provide stable spatial guidance during training, we employ teacher forcing by supplying ground-truth 2D-TGP as the condition for the TGP-Planner in training. While in test, to avoid the information leak, the 2D-TGP is first generated by the TGP-Prompter and subsequently fed into the TGP-Planner to produce the future trajectory. 

DriveTeach-VLA fully harnesses the intrinsic visual understanding ability of MLLMs and aligns it with driving-aware spatial reasoning for trajectory planning, achieving more precise trajectory prediction. Our main contributions are summarized as follows:

\begin{itemize}
    \item We identify that text-centric training in existing VLA models often produces spatially ungrounded attention.
    \item DriveTeach-VLA introduces DVD and 2D-TGP to explicitly teach VLAs {what to see} and {where to look}. For improving what to see, DVD transfers traffic visual cues via self-distillation from bbox-augmented images. For improving where to look, 2D-TGP projects expert trajectories onto the image plane to provide spatial guidance for planning.
    \item DriveTeach-VLA achieves the State-of-The-Art (SoTA) performance on {Navsim} and nuScene benchmarks.  
\end{itemize}

\section{Related Work}

\noindent\textbf{AD-VLA Model.} 
Recent AD-VLAs formulate trajectory prediction as a decoding task, where a MLLM planner generates a future few seconds' trajectory conditioned on visual–language context. The paradigm follows two major branches. The first relies on VLM itself's {autoregressive scheme}, where the model sequentially regresses the trajectory. ImpromptuVLA~\cite{chi2025impromptu}, AdaThinkDrive~\cite{luo2025adathinkdrive}, and other previous   models~\cite{hwang2024emma, xing2025openemma} regress text form trajectory directly. Poutine~\cite{rowe2025poutine} further explores a more suitable prompt and CoT reasoning steps for VLA's text  waypoint trajectory prediction. AutoVLA~\cite{zhou2025autovla} follows this scheme but instead replaces text-form trajectory with action tokens that discretize each dimension of the trajectory into several bins. The second relies on an additional planner to produce continuous trajectories conditioned on the latent reasoning features of the LLM. AsyncDriver~\cite{chen2024asynchronous}, RecogDrive~\cite{li2025recogdrive}, and Orion~\cite{fu2025orion} are built upon VLM's encoded features and introduce (query-based) MLP/Diffusion to predict the future trajectory. More recent works, such as DriveMoe~\cite{yang2025drivemoe} and DriveVLA-W0~\cite{li2025drivevlaw0}, employ a Mixture-of-Expert (MoE) architecture that pairs the VLM with an additional action-expert transformer by mutual attention to process multi-view images or reduce the inference time. The proposed DriveTeach-VLA follows the text-waypoint trajectory autoregressive scheme to explore the native capabilities of the MLLM backbone in the autonomous driving domain.

\noindent\textbf{AD-VLA Training.} Some AD-VLAs directly finetune  MLLMs for trajectory prediction~\cite{hwang2024emma, xing2025openemma}. To achieve stronger generalization, recent studies have converged toward a multi-stage training pipeline including:  \textit{(i) VQA Pretraining.}  VQA pairs are firstly pseudo-labeled by  MLLM, such as Qwen2.5-VL-72B and used to pretrain VLA with instruction learning. These VQA pairs involve multiple traffic-relevant tasks including scene description, traffic condition reasoning, and action explanation. DriveLM~\cite{sima2024drivelm} systematically organizes these tasks into a Perception–Prediction–Explanation–Behavior framework, which has since become the foundation for many subsequent VLA studies~\cite{renz2024carllava, luo2025adathinkdrive, zhou2025opendrivevla, li2025recogdrive}.  \textit{(ii) SFT-IL.} The model is supervised by expert trajectories and pseudo-labeled CoT reasoning steps with SFT to enable end-to-end trajectory generation from sensory inputs.  \textit{(iii) RL.}  AutoVLA~\cite{zhou2025autovla}, RecogDrive~\cite{li2025recogdrive}, and AdaThinkDrive~\cite{luo2025adathinkdrive} introduce GRPO-based~\cite{guo2025deepseek} reinforcement optimization. Especially, EVADrive~\cite{jiao2025evadrive} further proposes multi-objective RL framework, \textit{i.e.}, APO, for trajectory planning. The RL-stage aligns model prediction with human driving preference, including comfort, safety, and compliance with traffic regulations. DriveTeach-VLA replaces text-centric VQA pretraining stage with vision-centric DVD pretraining and augments IL and RL stages with 2D-TGP, enabling vision-grounded reasoning for reliable trajectory planning.

\section{Method}
\label{sec:method}
In this section, we first formulate the pipeline of the proposed method in Section~\ref{sec:formulation}. Then, we introduce DVD pretraining and TGP-Guided learning/inference in Section~\ref{sec:dvd} and Section~\ref{sec:tgp}, respectively.

\subsection{Preliminary}~\label{sec:formulation}

\noindent \textbf{BEV trajectory, 2D-TGP trajectory, and pinhole camera model.}
An expert trajectory with $T$ time-steps in world (BEV) coordinates is denoted as 
$\mathcal{T}_{w}=\{\mathbf{a}_t=(x_t, y_t, \psi_t ) \mid t=1,2,\ldots,T \}$, 
where $x_t$ and $y_t$ denote the position offsets in the ego-centric coordinate frame and $\psi_t$ denotes the yaw (heading) angle of the ego vehicle.
Given the front camera intrinsic matrix $K$ and extrinsic matrix $\left[ R \mid t \right]$, the $(x_t, y_t)$ in ego-centric coordinates can be projected onto the 2D image plane $(x_t^{I}, y_t^{I})$ using the pinhole camera model 
$\pi(\cdot)$~\cite{heyden2005multiple}:
\begin{equation}~\label{eq:bev-to-tgp}
\begin{aligned}
&(x_t^{I}, y_t^{I}) = \pi\!\left(K, [R \mid t], (x_t, y_t)\right), \mathcal{T}_{I} = \{\mathbf{a}^{I}_{t}=(x^I_t, y^I_t, \psi_t) \mid t=1,2,\ldots,T \}, \\
&P_I = \text{Text}\left[(x^I_1, y^I_1), (x^I_2, y^I_2), \ldots, (x^I_T, y^I_T)\right],
\end{aligned}
\end{equation}
where $P_I$ is named ground-truth 2D-TGP, and $\mathcal{T}_{I}$, combining the projected offset points with yaw angle, is named ground-truth 2D-TGP trajectory. {The yaw angle $\psi_t$ is retained from $\mathcal{T}_w$ so that the complete BEV trajectory can be recovered from the 2D-TGP trajectory via the inverse projection $\pi^{-1}$ (See Eq.~\ref{eq:tgp-to-bev} below).}

In general, the $\pi(\cdot)$ is not invertible due to the loss of depth information. 
However, in AD scenarios, trajectory points are typically assumed to lie on the ground plane $(z=0)$ and thus the projection becomes invertible:
\begin{equation}~\label{eq:tgp-to-bev}
(x_t, y_t) = \pi^{-1}\!\left(K, [R|t], (x_t^{I}, y_t^{I})\right) \quad \text{s.t.}\quad  z=0 ,
\end{equation}
which enables recovering a unique BEV position from the image-space trajectory and can be used to verify the quality of the estimated 2D-TGP trajectory ($\psi_t$ should be predicted separately for 2D-TGP trajectory evaluation). 

To establish a more intuitive understanding, we visualize the 2D-TGP trajectory on the image in Figure~\ref{fig:vis-case1}. By providing the MLLM with 2D-TGP coordinates, we can better align its intrinsic grounding ability with the VLA driving task.

\noindent \textbf{Task formulation.} Given observed camera images $C$, ego-state $S$, language instruction $L$, AD-VLA models output BEV trajectory estimation $\hat{\mathcal{T}}_w$ for planning:
\begin{equation}~\label{eq:AD-VLA-formulate}
    \hat{\mathcal{T}}_w = \text{AD-VLA}(C, S, L).
\end{equation}
Different from previous AD-VLAs formulating trajectory prediction as Eq.~\ref{eq:AD-VLA-formulate}, DriveTeach-VLA consists of two models, including TGP-Prompter and TGP-Planner, both of which are built upon Qwen2.5-VL-3B. TGP-Prompter firstly output 2D-TGP estimation $\hat{\mathcal{T}}_{I}$ as spatial guidance, serving as the textual condition for TGP-Planner to estimate BEV trajectory:
\begin{equation}~\label{eq:driveteach-vla-formulation}
    \hat{P}_I, \hat{\mathcal{T}}_I=\text{TGP-Prompter}(C, S, L_{\text{TGP}}), \quad \hat{\mathcal{T}}_{w} =\text{TGP-Planner}(C, S, [L; \hat{P}_I]),
\end{equation}
where $L_{\text{TGP}}$ and $L$ is the language instruction for TGP-Prompter and TGP-Planner, detailed in Supp. Figure~\ref{fig:sys_sft_tgp} and Supp. Figure~\ref{fig:sys_cot_rft}, respectively. 
Such textual coordinates conditions $\hat{P}_I$ are reasonable due to the intrinsic grounding ability of MLLMs, as 2D coordinates are naturally interpretable to the MLLM model. 

Furthermore, we decouple the TGP-Prompter and TGP-Planner into two separate models to avoid instruction confusion, as 2D-TGP generation and BEV trajectory prediction have distinct optimization objectives despite sharing similar input modalities.

\begin{figure*}[t]
    \centering
    \includegraphics[width=\linewidth]{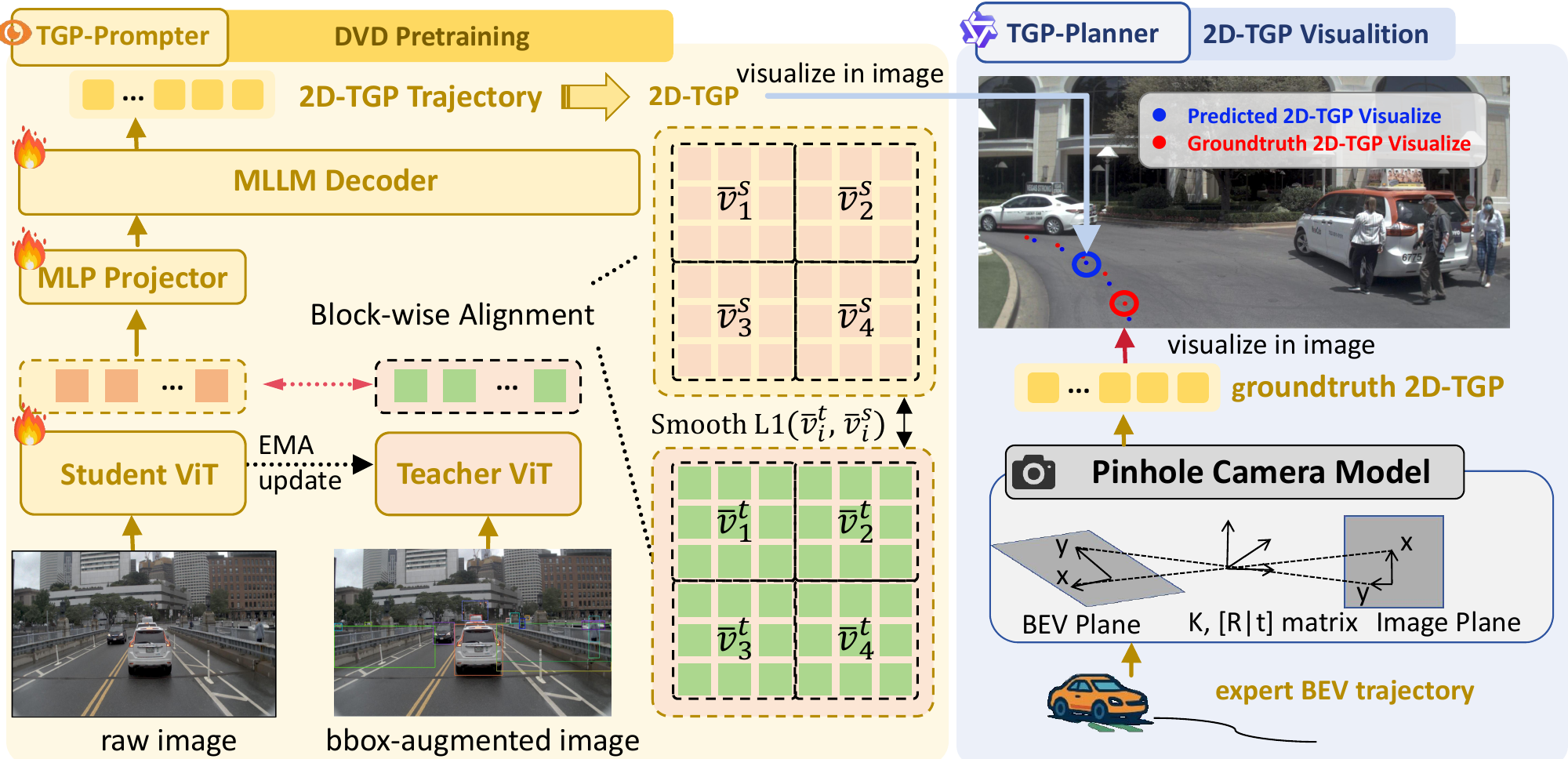}
    \caption{
\textbf{DriveTeach-VLA schemes.} \textcolor{orange!80!black}{\textbf{Left}}:  Traffic critical object priors are injected via bbox-augmented image self-distillation. \textcolor{blue!70!black}{\textbf{Right}}: The visualized 2D-TGP is highly related to the driving behavior (turn left), and the 2D-TGP conditions TGP-Planner in text form of a sequence of 2D coordinates, which is interpretable for MLLM.}
    \label{fig:dvd}
\end{figure*}

\subsection{TGP-Prompter: DVD Pretraining}~\label{sec:dvd}
DVD pretraining is specially designed for the TGP-Prompter to inject traffic priors without reliance on VQA pairs. Existing VLA models typically employ ViT encoders pretrained on natural images, which lack the domain knowledge about what should be seen during driving, \textit{e.g.}, lanes, traffic participants, and scene layout. To address this issue, DVD firstly introduces the bbox-augmented image self-distillation for the ViT encoder of the VLA model,   enabling it to learn driving-critical visual cues. In addition, to achieve precise driving-aware spatial perception regarding where to look, DVD pretraining also includes SFT that trains the VLA decoder to estimate the 2D-TGP trajectory. See Figure~\ref{fig:dvd} (left) for the overall training framework, and detailed elaborations are as follows:

\begin{figure*}[t]
    \centering
    \includegraphics[width=\linewidth]{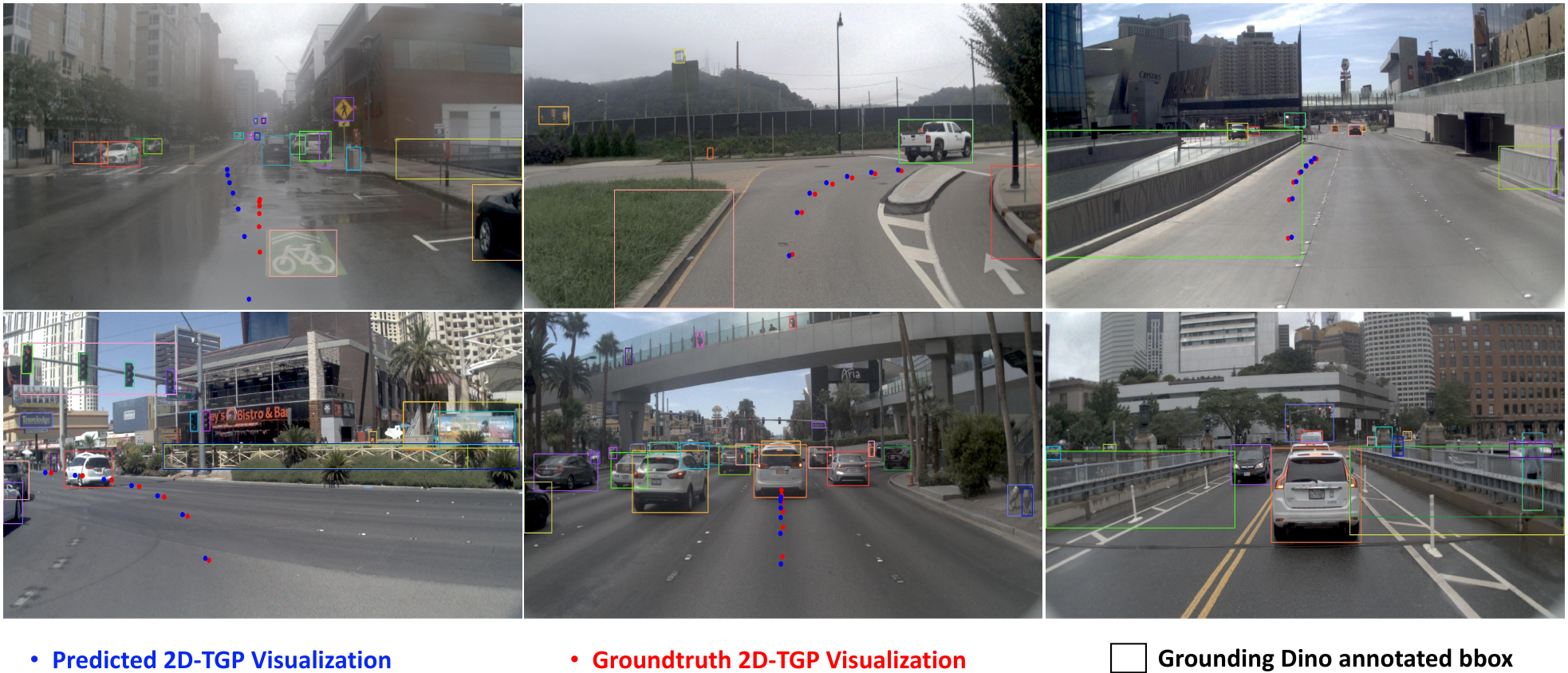}
    \caption{Visualization of the predicted 2D-TGP trajectory, critical objects detected by Grounding Dino of DriveTeach-VLA.}
    \label{fig:vis-case1}
\end{figure*}

\noindent\textbf{Bbox-augmented image self-distillation for ViT encoder.}  We firstly employ an open-vocabulary detector Grounding Dino~\cite{liu2024grounding} to detect traffic-critical objects, including \textit{car, truck, bus, trailer, construction vehicle, pedestrian, motorcycle, bicycle, barrier, traffic element, and traffic light}. The complete grounding object lists can be found in the Supp. Figure~\ref{fig:sys_dino}. By overlaying the detected bounding boxes on the raw camera image $C$, we can obtain the bbox-augmented image $C_{\text{bbox}}$. The bbox in Figure~\ref{fig:vis-case1} presents the object detection results. The self-distillation involves a pair of student and teacher ViT encoders, both initialized from the same MLLM's ViT encoder. The teacher ViT receives $C_{\text{bbox}}$ while the student ViT is fed with $C$:
\begin{equation}
\text{ViT}_t(C_{\text{bbox}})=\left[v_1^t,\cdots,v_N^t\right], \quad
\text{ViT}_s(C)=\left[v_1^s,\cdots,v_N^s\right],
\end{equation}
where  $N$ is the image patch token number, and $\text{ViT}_s$/$v_i^s$, and $\text{ViT}_t$/$v_i^t$ denotes student and teacher encoder/their patch features, respectively. In terms of feature alignment between  $v_i^t$ and $v_i^s$ , block-wise alignment is employed to capture broader contextual information brought by the bounding boxes. Specifically, we partition the feature maps into $K$ non-overlapping blocks $\mathcal{B}_k$ (\textit{i.e.}, dividing the feature map into a grid of rows $\times$ columns, e.g., $2\times4$ yields $K{=}8$ blocks) and compute block-wise alignment loss $\mathcal{L}_{\text{distill}}$:
\begin{equation}
\bar{v}_k^{t} = \frac{1}{|\mathcal{B}_k|}\sum_{i \in \mathcal{B}_k} v_i^{t}, \quad 
\bar{v}_k^{s} = \frac{1}{|\mathcal{B}_k|}\sum_{i \in \mathcal{B}_k} v_i^{s}, \quad \mathcal{L}_{\text{distill}} = 
\frac{1}{K}\sum_{k=1}^{K}\text{Smooth-L1}(\bar{v}_k^{t}, \bar{v}_k^{s}),
\end{equation}
where Smooth-L1 loss provides robustness to local spatial outliers~\cite{liu2021adaptive}.

\noindent\textbf{2D-TGP trajectory SFT for MLLM decoder.}  We first convert expert trajectories to 2D-TGP trajectories according to Eq.~\ref{eq:bev-to-tgp}. Then the TGP-Prompter is supervised to predict the 2D-TGP trajectory via optimizing the SFT cross-entropy loss:
\begin{equation}
\mathcal{L}_{{\text{TGP}}}
= -\frac1T \sum_{t=1}^T
\log P_\theta( \mathcal{T}_{I}^{(t)} \mid C, L_{\text{TGP}}, \mathcal{T}_{I}^{(<t)}),
\end{equation}
where $P(\cdot)$ is the TGP-Prompter policy, $L_{\text{TGP}}$ is language instruction (detailed in Supp. Figure~\ref{fig:sys_sft_tgp}), and $\mathcal{T}_{I}^{(<t)}$ denotes the trajectory action sequence before step $t$, and $\mathcal{T}_{I}^{(t)}$ denotes the action at step $t$. 

\noindent\textbf{Training Objective.}
The overall training loss is defined as:
\begin{equation}
\mathcal{L}_{\text{DVD}} = 
\mathcal{L}_{\text{distill}} + 
\lambda_{\text{TGP}} \cdot \mathcal{L}_{\text{TGP}},
\end{equation}
where  $\lambda_{\text{TGP}}$ is a hyperparameter. During training, the encoded features from the student ViT are fed into the MLLM decoder to make a prediction. The weights of the student ViT and the MLLM decoder are updated through backpropagation, while the teacher ViT is updated with Exponential Moving Average (EMA) from student ViT following the classic self-distillation setup from DINO~\cite{caron2021emerging}.

\subsection{TGP-Planner: TGP-guided Learning and Reasoning}~\label{sec:tgp}
The TGP-Planner is firstly initialized from the TGP-Prompter to inherit driving-related priors. It is then trained to generate step-by-step CoT reasoning and predict future trajectories in BEV space, conditioned on the 2D-TGP, the observed camera image, and the driving instruction. 
 Detailed processes are as follows.

\noindent\textbf{TGP-guided CoT Reasoning.}
Following Poutine~\cite{rowe2025poutine}, we generate pseudo-labeled reasoning data using the Qwen2.5-VL-72B model. The first three sub-tasks are identical to those in Poutine, while the fourth is modified to suit the 2D-TGP condition:

\textit{Task 1: Critical Object Detection.}
Determine whether any critical instance from a predefined object class list (\textit{e.g.}, hazards, cyclists, conflicting vehicles) may influence the ego vehicle’s future trajectory.

\textit{Task 2: Natural Language Explanation.}
Provide a concise explanation of the expert trajectory and the reasoning behind the expert driver’s behavior.

\textit{Task 3: Meta-Behavior Selection.}
Select a single behavior (\textit{e.g.}, turning, acceleration/deceleration, lane-following) that best describes the trajectory.

\textit{Task 4: Future Trajectory Prediction.} Different from prior work, we introduce spatial guidance via {2D-TGP}, where projected trajectory keypoints $\{(x_1^I,y_1^I),\dots,(x_8^I,y_8^I)\}$ indicate regions along the feasible driving path that the model should attend to. Task 4 is modified as: given the observed camera image, ego-state, reasoning outputs, and 2D-TGP keypoints, predict the optimal 4-second future trajectory.

Supp. Figure~\ref{fig:sys_cot_label} and Supp. Figure~\ref{fig:sys_cot_rft} details the prompts used for pseudo-labeling and TGP-Planner, respectively. We supervise TGP-Planner with above CoT reasoning data by SFT. After CoT-SFT training, we further align TGP-Planner's trajectory prediction with human driving preferences using reinforcement learning, following AutoVLA~\cite{zhou2025autovla}. At each rollout, the predicted trajectory $\hat{\mathcal{T}}_w$ is evaluated by the PDM-Score (PDMS)~\cite{dauner2024navsim}:

\begin{equation}~\label{eq:pdms}
\text{PDMS} = \prod_{c \in C} c \times \frac{\sum_{m \in M} w_m \cdot m}{\sum_{m \in M} w_m}, 
\end{equation}

where $C = \{\text{NC, DAC}\}$ (No-at-fault Collisions, Drivable Area Compliance) and $M = \{\text{EP, TTC, C}\}$ (Ego Progress, Time to Collision, Comfort) with weights $w_m = \{5, 5, 2\}$. We follow CuriousVLA~\cite{chen2026devil} to conduct data filtering and rollout for GRPO~\cite{guo2025deepseek}.

\noindent\textbf{Teacher forcing training and Inference pipeline.} During CoT-SFT and GRPO-RL, to produce the most precise guidance for driving skills learning, we used the ground-truth 2D-TGP to condition TGP-Planner. While in test (inference), as shown in Figure~\ref{fig:pipeline} (left), the  2D-TGP condition is firstly generated by TGP-Prompter, which is subsequently fed into TGP-Planner for future trajectory prediction.
Although this introduces a train–test gap, we show in Section~\ref{sec:ablation} that the gap is negligible ($\sim$0.4 PDMS), confirming TGP-Prompter provides sufficiently accurate spatial guidance.

\section{Experiment}

\begin{table*}[t!]
\centering
\small
\setlength{\tabcolsep}{1pt}
\caption{\textbf{PDMS Results on Navtest.} Best results within each category are in \textbf{bold}. $\ast$ Curious-VLA is fully open-sourced, and we reproduce its results with the same GRPO setting with us (1 round).}
\vspace{-20pt}
\label{tab:sota_navsim}
\begin{center}
\begin{tabular}{l|ccccccc|c}
\toprule
\textbf{Method} & \textbf{Pub.} & \textbf{Base VLM} & \textbf{NC } & \textbf{DAC} & \textbf{TTC } & \textbf{C } & \textbf{EP} & \textbf{PDMS } \\ 
\midrule
Human & & - & 100 & 100 & 100 & 99.9 & 87.5 & 94.8 \\ 
\midrule
\multicolumn{8}{l}{\textit{End-to-End Methods}} \\
UniAD~\cite{hu2023planning} & CVPR'23 & - & 97.8 & 91.9 & 92.9 & \textbf{100.0} & 78.8 & 83.4 \\
TransFuser~\cite{chitta2022transfuser} &TPAMI'22 & - & 97.7 & 92.8 & 92.8 & \textbf{100.0} & 79.2 & 84.0 \\ 
Hydra-MDP~\cite{li2024hydra} & CVPRW'24 & - & 98.3 & 96.0 & 94.6 & \textbf{100.0} & 78.7 & 86.5 \\
DiffusionDrive~\cite{liao2025diffusiondrive}& CVPR'25& - & 98.2 & 96.2 & 94.7 & \textbf{100.0} & 82.2 & 88.1 \\
WoTE~\cite{li2025end}& ICCV'25 & - & \textbf{98.5} & 96.8 & 94.4 & 99.9 & 81.9 & 88.3 \\
ASSCG~\cite{ang2026asscgjustrightgatingchattering} & Arxiv'26 & - & 98.2 &\textbf{ 98.3} & \textbf{94.8} & \textbf{100.0}&\textbf{87.5}&\textbf{91.4 }
\\
\midrule
\multicolumn{8}{l}{\textit{VLA-based Methods}} \\
ReCogDrive~\cite{li2025recogdrive}& ICLR'26 & InternVL3-8B & 98.2 & 97.8 & 95.2 & 99.8 & 83.5 & 89.6 \\
ImagiDrive-S~\cite{li2025imagidrive}& ICRA'26 & InternVL2.5-4B & 98.1 & 96.2 & 94.5 & \textbf{100.0} & 80.5 & 86.9 \\
AutoVLA~\cite{zhou2025autovla}& NeurIPS'25 & Qwen2.5-VL-3B & 98.4 & 95.6 & \textbf{98.0} & 99.9 & 81.9 & 89.1 \\
CuriousVLA$^{\ast}$~\cite{chen2026devil} & CVPR'26 & Qwen2.5-VL-3B & 97.7 & 95.9 & 97.2 & 98.2 & 89.2 & 88.9 \\
\rowcolor{gray!20}
DriveTeach-VLA & -& Qwen2.5-VL-3B & \textbf{98.5} & \textbf{96.9} & 97.9 & 98.2 & \textbf{88.5} &\textbf{ 90.4} \\ 
\bottomrule
\end{tabular}
\end{center}
\end{table*}

\begin{table}[t]
\centering

\small
\caption{\textbf{Open-loop trajectory prediction on nuScenes.} Best results within each category are in \textbf{bold}, second best are \underline{underlined}.}
\label{tab:nuscenes_l2}

\begin{tabular}{lccccc}
\toprule
\multirow{2}{*}{Method}  &\multirow{2}{*}{Pub.}& \multicolumn{2}{c}{ST-P3 metrics} & \multicolumn{2}{c}{UniAD metrics} \\
\cmidrule(lr){3-4} \cmidrule(lr){5-6}&& L2 (m) $\downarrow$ & Collision (\%) $\downarrow$ & L2 (m) $\downarrow$ & Collision (\%) $\downarrow$ \\
\midrule
UniAD~\cite{hu2023planning}  &CVPR'23& 0.69 & \textbf{0.12} & 1.03 & \textbf{0.31} \\
ST-P3~\cite{hu2022st}  &ECCV'22& 2.11 & 0.71 & -- & -- \\
VAD~\cite{jiang2023vad}   &ICCV'23& 0.37 & 0.14 & -- & -- \\
EMMA~\cite{hwang2024emma}  &TMLR'24& \underline{0.32} & -- & -- & -- \\
OpenEMMA~\cite{xing2025openemma}  &WCAC'25& 2.81 & -- & -- & -- \\
AutoVLA~\cite{zhou2025autovla}  &NeurIPS'25& 0.48 & \underline{0.13} & 0.86 & \underline{0.35} \\
Impromptu VLA~\cite{chi2025impromptu}  &NeurIPS'25& 0.33 & \underline{0.13} & \underline{0.67} & 0.38\\
\rowcolor{gray!20}
\textbf{DriveTeach-VLA }&-& \textbf{0.30} & \textbf{0.12} &\textbf{0.60} & \textbf{0.31} \\
\bottomrule
\end{tabular}
\end{table}
\subsection{Experimental Setup}
\noindent\textbf{Benchmarks.} We evaluate our method on two public autonomous-driving benchmarks: {non-reactive closed-loop} {NAVSIM}~\cite{dauner2024navsim} and {open-loop} {nuScenes}~\cite{qian2024nuscenes}.  
NAVSIM predicts future 4-second trajectory and is designed for intention-changing scenarios where the future motion of the ego vehicle cannot be reliably predicted from history alone. Following~RecogDrive\cite{li2025recogdrive}, we report PDMS results of DriveTeach-VLA on NAVSIM's navtest.  On nuScenes, following UniAD~\cite{hu2023planning} and ST-P3~\cite{hu2022st}, we report the average L2 trajectory error and collision rate over a 3-second horizon.

\noindent\textbf{Implementation Details.} We follow CuriousVLA~\cite{chen2026devil} to use the step-wise normalized text trajectory. Training uses AdamW (lr $4\times10^{-5}$, weight decay $0.05$) with a cosine schedule and 0.10 warm-up. Models are trained with batch size 16 on 8$\times$H100 GPUs. DriveTeach-VLA is trained in three stages.
\textit{(i) DVD pretraining}:1 epoch, with block size set to $ 2\times 4 $ and trajectory-loss weight $\lambda_{\text{TGP}}=0.1$. EMA momentum is set to $0.996$.
\textit{(ii) CoT-SFT}: 6 epochs using Chain-of-Thought annotations generated by Qwen2.5-VL-72B.  
\textit{(iii) GRPO-RL}: 180 update steps with a group size of 8. Especially, for fair comparison, CoT and GRPO are not employed for nuScene.

\subsection{Comparison with the SoTA models.}

\noindent \textbf{NAVSIM Results.} We benchmark against recently published end-to-end and VLA-based SoTA models, including UniAD, TransFuser, Hydra-MDP, DiffusionDrive, WoTE, RecogDrive, ImagiDrive-S, and AutoVLA. Among them, AutoVLA is the most similar to DriveTeach-VLA, as both adopt an autoregressive trajectory prediction paradigm. We report results on the Navtest split using  PDMS metrics, and results are exhibited in Table~\ref{tab:sota_navsim}. DriveTeach-VLA achieves 90.4 PDMS on Navtest with a single inference pass.

\noindent\textbf{NuScenes Results.} Table~\ref{tab:nuscenes_l2} shows the comparison of DriveTeach-VLA with several SOTA methods using the L2 distance error (L2) and collision rate as the primary metrics.  DriveTeach-VLA achieves the best performance on both metrics, with an L2 of 0.30 and a collision rate of 0.12, surpassing all prior methods. The model exhibits the highest precision in trajectory prediction with minimal collision occurrence, indicating superior performance in real-world driving scenarios.

\subsection{Ablation and Analysis}~\label{sec:ablation}

\noindent\textbf{Attention Analysis of Vision Encoders.} We compare the attention maps of the TGP-Planner during trajectory decoding with a CoT-SFT trained Qwen2.5-VL-3B baseline. The qualitative observations in Figure~\ref{fig:vit_attention_decode} show that the baseline exhibits dispersed attention, while DriveTeach-VLA produces clear spatial grounding on traffic-critical objects. To quantify this phenomenon, we introduce a metric called \textit{Attention Mass (AM)}. The AM score for an observed camera image $C$ is calculated as follows. Firstly, we compute the attention maps between the decoded waypoint tokens and the visual patch tokens. The attention maps are averaged across all waypoint tokens to obtain an aggregated attention map $A$. Then, let $\mathcal{R}$ denote the set of bounding boxes obtained from the bbox-augmented image $C_{\text{bbox}}$ (Section~\ref{sec:dvd}), and let $|r|$ denote the area of box $r$. We sum the attention values of visual tokens lying inside all bounding boxes and normalize by the total box area:
\begin{equation}
\label{eq:attn_mass}
\mathrm{AM}(C)
= \frac{1}{\sum_{r\in\mathcal{R}} |r|}
\sum_{r\in\mathcal{R}}
\sum_{(i,j)\in r} A_{i,j},
\end{equation}
where $A_{i,j}$ denotes the average attention value at spatial position $(i,j)$ in the averaged attention map. We randomly sample 1k scenes from Navtrain and compute the mean and variance of their AM scores. Table~\ref{tab:vit_attention_stats} reports the results. 
The CoT-SFT baseline exhibits significantly lower AM than the TGP-Planner and also shows a smaller variance, indicating consistently weak attention on traffic-critical objects. This verifies that DVD pretraining effectively injects critical traffic-object priors into DriveTeach-VLA's TGP-Planner.

\begin{table}[t]
\centering

\begin{minipage}{0.48\linewidth}
\centering
\caption{\textbf{Ablation on  $\lambda_{\text{TGP}}$ with the TGP-Prompter.} A moderate weight yields the best results.}
\label{tab:ablation_lambda}
\small
\begin{tabular}{cccccc}
\toprule
$\lambda_{\text{TGP}}$ & NC  & DAC  & TTC  & EP & PDMS  \\
\midrule
0.05 & 97.5 & 94.5 & 96.7 & 86.9 & 86.3 \\
0.08 & 97.5 & 94.3 & 96.3 & 87.0 & 86.1 \\
\textbf{0.10} & \textbf{98.0} & \textbf{95.5} & \textbf{97.1} & \textbf{87.1} & \textbf{87.6} \\
0.12 & 97.8 & 95.1 & 97.0 & 86.9 & 87.2 \\
0.15 & 97.9 & 95.2 & \textbf{97.1} & 86.9 & 87.4 \\
\bottomrule
\end{tabular}
\end{minipage}
\hfill
\begin{minipage}{0.48\linewidth}
\centering
\caption{\textbf{Ablation on  $\mathcal{B}_k$ with the TGP-Prompter.} Both patch-/block-wise DVD improve performance.}
\label{tab:ablation_block}
\small
\begin{tabular}{lcccc}
\toprule
$\mathcal{B}_k$  & NC  & DAC  & TTC & PDMS  \\
\midrule
w/o DVD & - & - & - & 86.2 \\ \midrule

2$\times$4 & \textbf{98.0} & \textbf{95.5} & \textbf{97.1} & \textbf{87.6} \\
4$\times$7 & 97.7 & 95.1 & 96.9 & 87.1 \\
Patch-by-patch & 97.7 & 94.5 & 96.7 & 86.5 \\
\bottomrule
\end{tabular}
\end{minipage}

\end{table}

\begin{table}[t]
\centering
\caption{\textbf{Ablation of DVD pretraining and 2D-TGP on Navsim.} $\dagger$ indicates we convert 2D-TGP trajectory generated by TGP-Prompter to BEV trajectory with Eq.~\ref{eq:tgp-to-bev}.}
\label{tab:ablation_dvd_tgp}
\begin{tabular}{l|l|cccc}
\toprule
Method & Model &DAC  & EP & TTC  & PDMS  \\
\midrule
QwenVL-2.5-3B & Planner & 93.2 & 85.8 & 97.3 & 84.8 \\
 + VQA + CoT & Planner & 94.0 & 87.2 & 96.7 & 86.4 \\ \midrule
+ DVD + CoT & Prompter $^{\dagger}$  & 94.9 & {87.5} & {96.7} & 87.3 \\ 
+ DVD + CoT+GRPO & Prompter $^{\dagger}$  & 95.5 & 90.2 & 96.5 & 88.2 \\ \midrule
+ DVD + CoT & Planner& {94.4} & 86.3 & 97.8 & 87.1 \\
+ DVD + TGP + CoT & Planner& {95.8} & 87.2 & 96.6 &  {88.2} \\
+ DVD +  CoT +GRPO & Planner& 95.8 & \textbf{90.6} & 96.7 & {89.1} \\
+ DVD + TGP + CoT +GRPO & Planner& \textbf{96.9} & 88.5 & \textbf{97.9} & \textbf{90.4} \\
\bottomrule
\end{tabular}
\end{table}

\begin{table}[t]
\centering

\begin{minipage}[t]{0.46\linewidth}
\centering

\caption{\textbf{Attention Mass Statistics.}
AM-Score computed on 1k randomly sampled Navtrain scenes.}
\label{tab:vit_attention_stats}

\footnotesize
\begin{tabular}{lcc}
\toprule
AM-Score & Mean ($\times 10^{-2}$) & Std ($\times 10^{-2}$) \\
\midrule
Qwen2.5-VL-3B & 2.28 & 3.12 \\
TGP-Planner & 3.19 & 3.77 \\
\bottomrule
\end{tabular}

\end{minipage}
\hfill
\begin{minipage}[t]{0.45\linewidth}
\centering

\caption{\textbf{DVD robustness under detector noise.}
PDM-Score under random bbox drop and jitter perturbations.}
\label{tab:dvd-robustness}

\footnotesize
\begin{tabular}{lcccc}
\toprule
Perturb. & w/o DVD & 0\% & 20\% & 40\% \\
\midrule
Random drop & 86.2 & 87.6 & 86.8 & 85.3 \\
Jitter bbox      & 86.2 & 87.6 & 87.1 & 86.6 \\
\bottomrule
\end{tabular}

\end{minipage}

\end{table}

\noindent \textbf{Hyperparameter analysis of  $\lambda_{\text{TGP}}$ and $\mathcal{B}_k$}
As this evaluation only involves DVD pretraining, \textit{i.e.}, TGP-Prompter only, we transform the predicted 2D-TGP trajectories into BEV trajectories using Eq.~\ref{eq:tgp-to-bev} and calculate PDMS on Navtest. Table~\ref{tab:ablation_lambda} studies the hyperparameter $\lambda_{\text{TGP}}$ balancing 2D-TGP learning and self-distillation in DVD pretraining. A small value ($0.05$) under-utilizes trajectory cues, while a large value ($1.5$) overemphasizes the auxiliary prediction task. A moderate value ($0.1$) achieves the best performance across NC, DAC, TTC, and PDMS.

Table~\ref{tab:ablation_block} analyzes how the block participation strategy $\mathcal{B}_k$ in DVD affects planning quality.
The 2$\times$4 partition achieves the best PDMS among the tested configurations. This is likely because DVD relies on bbox-based visual prompts highlighting traffic-critical objects, and blocks with a larger receptive field during distillation better capture these highlighted regions and their surrounding context. In contrast, patch-level alignment dilutes the signal since most patches fall outside the highlighted bounding boxes.

\noindent \textbf{Ablation on TGP-guided CoT and DVD.} The incremental ablation results are reported in Table~\ref{tab:ablation_dvd_tgp}. Ablation begins based on a single Qwen2.5-VL-3B model (84.8 PDMS). We first establish a baseline using conventional VQA pretraining followed by CoT-SFT training (second row, 86.4 PDMS). 
Next, we introduce DVD pretraining combined with CoT-SFT. 
Under this setting, only the TGP-Prompter is trained: after completing DVD pretraining, the TGP-Planner model is initialized from TGP-Prompter and further optimized with CoT-SFT (87.1 PDMS). Subsequently, 2D-TGP guided reasoning is employed and brings performance to 88.2. Finally, the GRPO boosts performance to 90.4. 

Furthermore, Table~\ref{tab:ablation_dvd_tgp}'s 3-4 rows isolate the performance of TGP-Prompter, and 5-8 rows ablate TGP-guided CoT reasoning with and without GRPO.

\begin{table}[t]
  \centering
  \caption{Efficiency and Dual-model Design Ablation on H100. Only DVD and TGP-guided CoT-SFT is used. $\dagger$ indicates we convert 2D-TGP trajectory generated by TGP-Prompter to BEV trajectory with Eq.~\ref{eq:tgp-to-bev}. Type column indicates the number of used model.}
  \label{tab:efficiency-pdms}
  \setlength{\tabcolsep}{6pt}
  \begin{tabular}{@{}llcccc@{}}
    \toprule
    \textbf{Method} & \textbf{Setting} & \textbf{Type} & \textbf{Latency (s)} & \textbf{Mem. (GiB)} & \textbf{PDMS} \\
    \midrule
    DriveTeach & Prompter-Planner & dual & 3.03 & 17.2 & 88.2 \\
    Call 1 & Prompter & -- & 1.14 & 8.2 & -- \\
    Call 2 & Planner & -- & 1.89 & 9.0 & -- \\
    \midrule
    DriveTeach & Multi-turn QA & single & 2.87 & 8.6 & 87.0 \\
    Turn 1 & Prompter QA & -- & 1.11 & -- & -- \\
    Turn 2 & Planner QA & -- & 1.75 & -- & -- \\
    \midrule
    Baseline & Prompter $^{\dagger}$& single & 1.14 & 8.2 & 87.3 \\
    \bottomrule
  \end{tabular}
\end{table}

\noindent{\textbf{Train-Test Gap Analysis of the Predicted 2D-TGP.}
The teacher-forcing strategy uses ground-truth 2D-TGP during training but predicts $\hat{P}_I$ during inference. To quantify this train–test gap, we compare the standard inference pipeline against an oracle setting that leaks ground-truth 2D-TGP at test time.
As shown in Table~\ref{tab:tgp_gap}, leaking ground-truth 2D-TGP only increases PDMS by 0.4 (90.4 $\rightarrow$ 90.8), confirming that the TGP-Prompter produces reliable spatial guidance and the teacher-forcing gap is not a performance bottleneck.

Furthermore, we analyze the effect of 2D-TGP on TGP-Planner's prediction quality. Following Table~\ref{tab:tgp_gap}, we linearly bin samples by predicted $\hat{\mathcal{T}}_I$-vs-GT ${\mathcal{T}}_I$'s L2 error and report PDMS in Figure~\ref{fig:tgp_gap_analysis}. The small oracle gap (Table~\ref{tab:tgp_gap}: 90.4/90.8) is expected: most predicted TGPs are accurate. And when the error grows, PDMS drops gradually, showing robustness rather than TGP irrelevance.

\begin{table}[t]
\centering

\begin{minipage}[t]{0.52\linewidth}
\centering
\caption{\textbf{Train–test gap analysis of 2D-TGP.} Using predicted $\hat{P}_I$ at inference yields comparable performance to the oracle ground-truth $P_I$ setting.}
\label{tab:tgp_gap}
\small
\begin{tabular}{lcccc}
\toprule
Test-time 2D-TGP & DAC  & EP & TTC  & PDMS \\
\midrule
Predicted $\hat{P}_I$ & \textbf{96.9} & \textbf{88.5} & 97.9 & 90.4 \\
Ground-truth $P_I$& 96.5 & 87.5 & \textbf{99.3} & \textbf{90.8} \\
\bottomrule
\end{tabular}
\end{minipage}
\hfill
\begin{minipage}[t]{0.43\linewidth}
\centering
\caption{\textbf{Runtime comparison (seconds per sample) on L20 GPUs.} AutoVLA's runtime is sourced from its paper~\cite{zhou2025autovla}. }
\small
\renewcommand{\arraystretch}{0.8}
\setlength{\tabcolsep}{3pt} 
\label{tab:efficiency}
\begin{tabular}{lrr}
\toprule
Method & Trajectory & Runtime  \\
\midrule
ours & text waypoint & 3.18 s \\
AutoVLA & physical action & 3.95 s \\
AutoVLA & text waypoint & 7.65 s \\
\bottomrule
\end{tabular}
\end{minipage}

\end{table}

\begin{figure}[t]
\centering
\includegraphics[width=\linewidth]{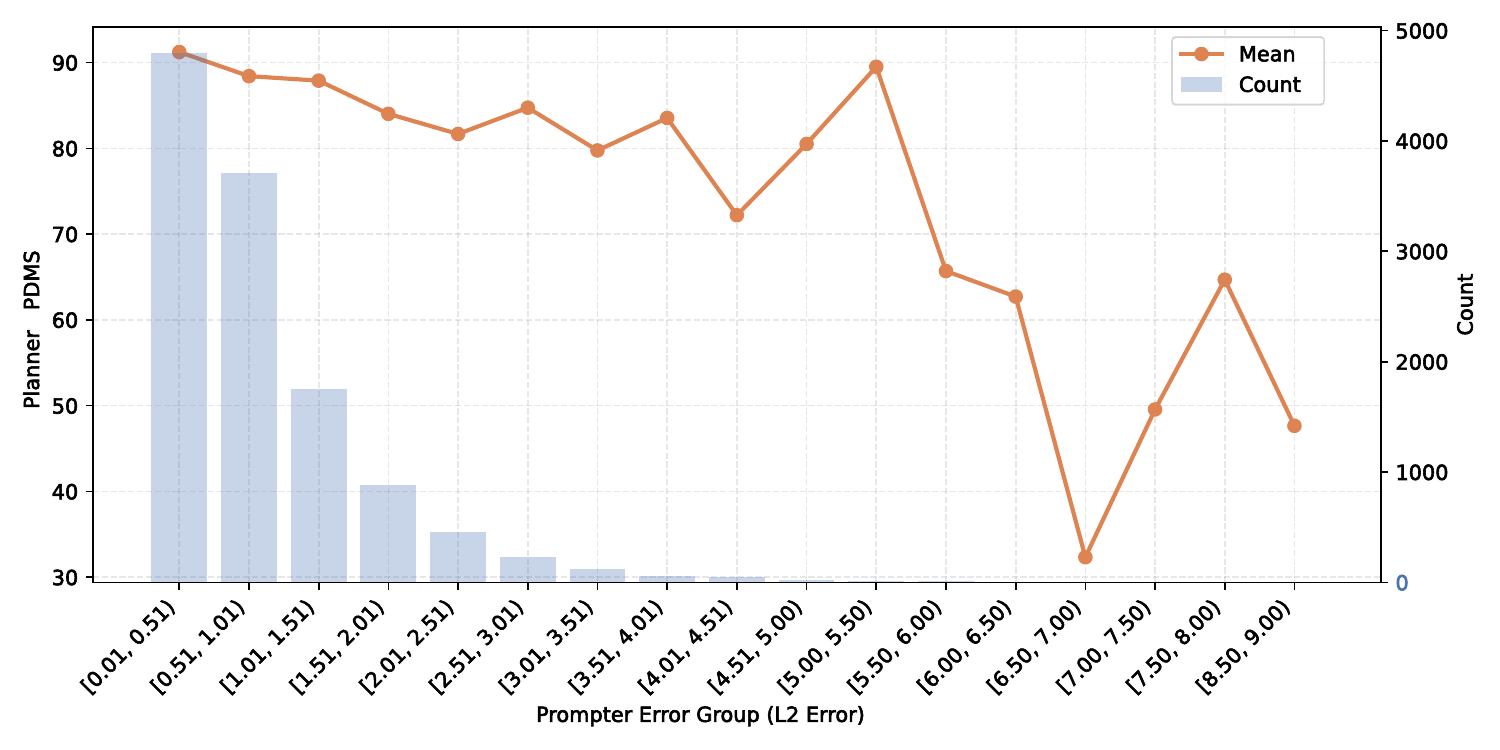}
\caption{\textbf{PDMS by predicted-vs-GT 2D-TGP L2 error.}
PDMS remains stable across a broad range of trajectory prediction errors, indicating that the policy is robust to moderate inaccuracies in 2D-TGP prediction.}
\label{fig:tgp_gap_analysis}
\end{figure}

\noindent{\textbf{Robustness of DVD.}} DVD depends heavily on Grounding DINO detection quality. If Grounding DINO misses critical objects in novel scenes or domain-shifted data (\textit{e.g.}, unusual vehicles, occluded pedestrians), the distilled visual priors may be noisy or misleading. So Table~\ref{tab:dvd-robustness} adds random bbox drop and jitter bbox noise on the annotated bboxes and tests TGP-Prompter's PDM-Score.  
The noise strength denotes the dropped-box ratio or the corner perturbation ratio w.r.t. box width/height. DVD shows robustness and remains helpful under moderate noise (20\% drop/jitter: 86.8/87.1 vs. w/o DVD 86.2), while severe 40\% noise can mislead guidance and hurt.

\noindent \textbf{Efficiency and Dual Model Design Ablation.} Although DriveTeach-VLA achieves SoTA performance, the dual-model architecture increases runtime latency. As shown in Table~\ref{tab:efficiency-pdms}, Prompter-Planner design increases 1x runtime (about 1.14s on H100) during inference. To assess its necessity, we replace it with a multi-turn QA variant. As shown in Table~\ref{tab:efficiency-pdms}, without GRPO, the Prompter-Planner design improves PDMS over the Prompter-only baseline (88.2 vs. 87.3), while the multi-turn QA variant performs worse (87.0  vs. 87.3). This suggests that jointly learning two instruction-following behaviors in one model remains challenging. 

Despite the longer runtime, we also found something surprised. We compare DriveTeach-VLA with its most similar baseline, AutoVLA. Table~\ref{tab:efficiency} reports the runtime results, measured on one L20 GPU. AutoVLA reports results under two settings: predicting text waypoint  and  action token trajectories. Although action tokens require fewer tokens and are therefore more efficient, AutoVLA relies on long CoT reasoning, which introduces substantial token overhead. Consequently, even its action-token variant remains slower than DriveTeach-VLA. This indicates that DriveTeach-VLA uses fewer CoT reasoning tokens while still outperforming AutoVLA, highlighting the efficiency advantage of our vision-grounded spatial enhancement.

\section{Conclusion}
This work identifies a key limitation of existing VLA paradigms: their heavy reliance on text-centric pretraining, which leads to trajectory decoding without explicit semantic guidance. To address this issue, we propose DriveTeach-VLA, a VLA framework that decouples spatial guidance from trajectory planning through a dual-module architecture consisting of a TGP-Prompter and a TGP-Planner. Extensive experiments on NAVSIM and nuScenes demonstrate the effectiveness of the proposed approach, supported by comprehensive ablation studies. Furthermore, despite the dual-model design increases runtime, our efficiency analysis shows that visual spatial enhancement reduces the reliance on reasoning tokens. These results highlight the importance of vision-grounded spatial guidance for improving  performance  in autonomous driving systems.

\section{Acknowledge}

DriveTeach-VLA is funded by the National Key Research and Development Program of China, 2024YFE0217600, Xiongan AI Institute, DiDi and Wuxi Research Institute of Applied Technologies, Tsinghua University under Grant 20242001120.

\clearpage

\bibliographystyle{splncs04}
\bibliography{main}

\clearpage
\setcounter{page}{1}
\appendix

\section{NAVSIM Best-of-N PDMS \& EPDMS}
Autoregressive model often exhibit strong exploration capability. In Table~\ref{tab:sota_navsim_bst_of_n}, we compare the best-of-N performance. For example, AutoVLA reports a best-of-N (BoN=6) PDMS of 92.1~\cite{zhou2025autovla}. But BoN strategy leaks test-set information and violates the NAVSIM evaluation rule. DrivorR~\cite{kirby2026driving} proposes an alternative approach by training a selector model to identify the best trajectory. Following this strategy, we perform N=12 times rollout for each scene and select the best trajectory using the DrivorR selector. This yields  92.7 PDMS, surpassing prior methods by a large margin.

\begin{table*}[ht]
\centering
\small
\setlength{\tabcolsep}{1pt}
\caption{\textbf{PDMS Results on Navtest.} \dag: Rollout N=12 trajectory and use DrivorR~\cite{kirby2026driving}'s selector to pick out the best for evaluation.}
\label{tab:sota_navsim_bst_of_n}
\vspace{-20pt}
\begin{center}
\begin{tabular}{l|ccccccc|c}
\toprule
\textbf{Method} & \textbf{Pub.} & \textbf{Base VLM} & \textbf{NC } & \textbf{DAC} & \textbf{TTC } & \textbf{C } & \textbf{EP} & \textbf{PDMS } \\ 
\midrule
Human & & - & 100 & 100 & 100 & 99.9 & 87.5 & 94.8 \\ 
\midrule
\multicolumn{8}{l}{\textit{VLA-based Methods}} \\
ReCogDrive~\cite{li2025recogdrive}& ICLR'26 & InternVL3-8B & 98.2 & 97.8 & 95.2 & 99.8 & 83.5 & 89.6 \\
ImagiDrive-S~\cite{li2025imagidrive}& ICRA'26 & InternVL2.5-4B & 98.1 & 96.2 & 94.5 & \textbf{100.0} & 80.5 & 86.9 \\
AutoVLA~\cite{zhou2025autovla}& NeurIPS'25 & Qwen2.5-VL-3B & 98.4 & 95.6 & {98.0} & 99.9 & 81.9 & 89.1 \\
DriveVLA-W0~\cite{li2025drivevlaw0} & ICLR'26 & Emu3-8B & 98.7 & \textbf{99.1} & 95.3 & 99.3 & 83.3 & 90.2 \\ 
AutoVLA  \dag  &  NeurIPS'25 & Qwen2.5-VL-3B & 99.1 & 98.8 & 87.9 & 97.2 & 100.0 & 92.1 \\
\rowcolor{gray!20}
DriveTeach-VLA & -& Qwen2.5-VL-3B & 98.5 & 96.9 & 97.9 & 98.2 & \textbf{88.5} & 90.4 \\ 

\rowcolor{gray!20}
\textbf{DriveTeach-VLA \dag }&- & Qwen2.5-VL-3B & \textbf{99.8} & {98.3} & \textbf{98.3} & 99.9 & {86.6} & \textbf{92.7} \\

\bottomrule
\end{tabular}
\end{center}
\end{table*}
\vspace{-30pt}
\begin{table*}[ht]
\footnotesize
\centering
\caption{\textbf{Extended PDMS (EPDMS) results on Navtest. } \dag: Rollout N=12 trajectory and use DrivorR~\cite{kirby2026driving}'s selector to pick out the best for evaluation. }
\label{tab:main_epdms}
\begin{tabular}{l|ccccccccc|c}
\toprule
Method & NC  & DAC  & DDC  & TLC  & EP  & TTC  & LK  & C  & EC  & \textbf{EPDMS} $\uparrow$ \\
\midrule
TransFuser~\cite{chitta2022transfuser} & 96.9 & 89.9 & 97.8 & 99.7 & 87.1 & 95.4 & 92.7 & 98.3 & 87.2 & 76.7 \\
HydraMDP++~\cite{li2025hydra} & 97.2 & 97.5 & 99.4 & 99.6 & 83.1 & 96.5 & 94.4 & 98.2 & 70.9 & 81.4 \\
ARTEMIS~\cite{feng2025artemis} & 98.3 & 95.1 & 98.6 & 99.8 & 81.5 & 97.4 & 96.5 & 98.3 & -- & 83.1 \\
ReCogDrive~\cite{li2025recogdrive} & 98.3 & 95.2 & {99.5}& 99.8 & 87.1 & 97.5 & 96.6 & 98.3 & 86.5 & 83.6 \\
DiffusionDrive~\cite{liao2025diffusiondrive} & 98.2 & 95.9 & 99.4 & 99.8 & 87.5 & 97.3 & 96.8 & 98.3 & 87.7 & 84.5 \\
\rowcolor{gray!20}
{DriveTeach-VLA} & {98.5} &  {96.9} & 99.2 & 99.8 & {88.5} &  {97.9} &  {96.9} & 98.2 & 81.2 & {85.4} \\
\rowcolor{gray!20}
\textbf{DriveTeach-VLA} \dag & \textbf{99.8} & \textbf{98.3} & \textbf{99.8}& 99.8 & \textbf{88.7} & \textbf{99.5} & \textbf{98.2} & 98.1 & 88.3 & \textbf{89.0} \\
\bottomrule
\end{tabular}
\end{table*}
\clearpage
\section{NuScenes Results}

\begin{table*}[ht]
  \caption{Open-loop trajectory prediction on the nuScenes dataset with more metrics  (where \textsuperscript{1} indicates sourced from \cite{qiao2025lightemma} Best results within each category are in \textbf{bold}, second best are \underline{underlined}. }
  \tiny
  \label{tab:nuscenes_quantitative}
  \centering
  \resizebox{\textwidth}{!}{
  \setlength{\tabcolsep}{3pt}
  \begin{tabular}{lcccccccccccccccc}
    \toprule
    \multirow{3}{*}{Method} 
    & \multicolumn{8}{c}{\textbf{ST-P3 metrics}} 
    & \multicolumn{8}{c}{\textbf{UniAD metrics}} \\
    \cmidrule(lr){2-9} \cmidrule(lr){10-17}
    & \multicolumn{4}{c}{L2 (m) ↓} 
    & \multicolumn{4}{c}{Collision (\%) ↓} 
    & \multicolumn{4}{c}{L2 (m) ↓} 
    & \multicolumn{4}{c}{Collision (\%) ↓} \\
    \cmidrule(lr){2-5} \cmidrule(lr){6-9} \cmidrule(lr){10-13} \cmidrule(lr){14-17}
    & 1s & 2s & 3s & Avg. 
    & 1s & 2s & 3s & Avg. 
    & 1s & 2s & 3s & Avg. 
    & 1s & 2s & 3s & Avg.\\
    \midrule

    ST-P3\textsuperscript{1}\cite{hu2022st} & 1.33 & 2.11 & 2.90 & 2.11 & 0.23 & 0.62 & 1.27 & 0.71 & -& -& -& -& -& -& -& -\\
    VAD\textsuperscript{1}\cite{jiang2023vad}  & 0.17 & 0.34 & 0.60 & 0.37 & 0.07 & 0.10 & 0.24 & 0.14  & -& -& -& -& -& -& -& - \\
    UniAD\textsuperscript{1}\cite{hu2023planning} & 0.44 & 0.67 & 0.96 & 0.69 & 0.04 & 0.08 & 0.23 & 0.12 & 0.48 & 0.96 & 1.65 & 1.03 & 0.05 & \textbf{0.17} & 0.71 & \underline{0.31}\\
    EMMA\textsuperscript{1}\cite{hwang2024emma}& \underline{0.14} & \underline{0.29} & \underline{0.54} & \underline{0.32} & -& -& -& - & -& -& -& -& -& -& -& - \\
    OpenEMMA\textsuperscript{1}\cite{xing2025openemma} & 1.45 & 3.21 & 3.76 & 2.81 & -& -& -& - & -& -& -& -& -& -& -& - \\
    OpenDriveVLA-3B\textsuperscript{1}\cite{zhou2025opendrivevla} & \underline{0.14} & 0.30 & 0.55 & 0.33 & \underline{0.02} & \underline{0.07} & \underline{0.22} & \textbf{0.10} & \underline{0.19} & \underline{0.58} & 1.24 & \underline{0.67} & \textbf{0.02} & \underline{0.18} & \underline{0.70} & \textbf{0.30}\\
    AutoVLA\textsuperscript{1}~\cite{zhou2025autovla} & 0.21 & 0.38 & 0.60 & 0.40 & 0.13 & 0.18 & 0.28 & 0.20  & 0.28 & 0.66 & \textbf{1.16} & 0.70 & 0.14 & 0.25 & \textbf{0.53} & 0.31  
    \\
    DriveTeach-VLA-3B & \textbf{0.13} & \textbf{0.27} & \textbf{0.51} & \textbf{0.30} & \textbf{0.01} & \textbf{0.09} & \textbf{0.25} & \underline{0.12} & \textbf{0.17} & \textbf{0.51} & \underline{1.20} & \textbf{0.60} & \textbf{0.02} & 0.21 & \underline{0.70} & \underline{0.31} 
    \\ 
    \bottomrule
  \end{tabular}
  }
\end{table*}

\begin{table*}[ht]
\centering
\caption{Open-loop trajectory prediction L2 errors (m) on the nuScenes dataset. (where \textsuperscript{1} indicates sourced from \cite{qiao2025lightemma}, \textsuperscript{2} indicates sourced from \cite{xing2025openemma} and \textsuperscript{3} indicates sourced from \cite{hwang2024emma}). \textsuperscript{4} indicates sourced from \cite{chi2025impromptu}. Best results within each category are in \textbf{bold}, second best are \underline{underlined}.}
\label{table:finetune-L2}
\begin{tabular}{lccccc}
\toprule
\multirow{2}{*}{\textbf{Method}} & \multicolumn{4}{c}{\textbf{L2 Error (m) $\downarrow$}} \\
\cmidrule(lr){2-5}
& 1s & 2s & 3s & \textbf{Avg.} \\
\midrule
\multicolumn{5}{l}{\textit{Generalist VLMs}}  \\
\midrule
GPT-4o\textsuperscript{1} \cite{hurst2024gpt} & \textbf{0.28} & \textbf{0.93} & \textbf{2.02} & \textbf{1.07} \\
Claude-3.7-Sonnet\textsuperscript{1} & \textbf{0.28} & \underline{0.94} & \underline{2.04} & \underline{1.09} \\
Gemini-2.5-Pro\textsuperscript{1} & 0.37 & 1.35 & 2.96 & 1.56 \\
LLaVA-1.6-Mistral-7B\textsuperscript{2} & 1.49 & 3.38 & 4.09 & 2.98 \\
Qwen2-VL-7B-Instruct\textsuperscript{2} \cite{Qwen2.5} & 1.45 & 3.21 & 3.76 & 2.81 \\
DeepSeek-VL2-16B\textsuperscript{1} \cite{DeepSeek-VL2} & 0.66 & 1.68 & 2.92 & 1.75 \\
LLaMA-3.2-11B-Vision-Instruct\textsuperscript{1} & 0.52 & 1.42 & \underline{2.68} & \underline{1.54} \\
Qwen-2.5-VL-7B-Instruct\textsuperscript{1} \cite{Qwen2.5} & \underline{0.46} & \textbf{1.33} & \textbf{2.55} & \textbf{1.45} \\
\midrule
\multicolumn{5}{l}{\textit{Training-based Driving Specialists (Existing Methods)}} \\
\midrule
UniAD\textsuperscript{3} \cite{hu2023planning} & 0.42 & 0.64 & 0.91 & 0.66 \\
VAD\textsuperscript{3} \cite{jiang2023vad} & 0.17 & 0.34 & 0.60 & 0.37 \\
BEV-Planner\textsuperscript{3} \cite{li2024ego} & \underline{0.16} & \textbf{0.32} & \textbf{0.57} & \textbf{0.35} \\
Ego-MLP\textsuperscript{3}* \cite{li2024ego} & \textbf{0.15} & \textbf{0.32} & \underline{0.59} & \textbf{0.35} \\
\midrule
\multicolumn{5}{l}{\textit{Ours and Key Competitors (Specialized Driving Models)}} \\
\midrule
DriveVLM\textsuperscript{3} \cite{tian2024drivevlm} & 0.18 & 0.34 & 0.68 & 0.40 \\
OmniDrive\textsuperscript{3} \cite{wang2024omnidrive} & {0.14} & 0.29 & 0.55 & 0.33 \\
DriveVLM-Dual\textsuperscript{3} \cite{tian2024drivevlm} & 0.15 & 0.29 & \textbf{0.48} & \underline{0.31} \\
EMMA (random init) \cite{hwang2024emma}\textsuperscript{3} & 0.15 & 0.33 & 0.63 & 0.37 \\
EMMA \cite{hwang2024emma}\textsuperscript{3} & {0.14} & \underline{0.29} & 0.54 & 0.32 \\
Impromptu-VLA-3B \textsuperscript{4} & \underline{0.13} & \textbf{0.27} & {0.52} & \underline{0.31} \\
Impromptu-VLA-7B  \textsuperscript{4} & \underline{0.13} & \textbf{0.27} & 0.53 & \underline{0.31} \\
DriveTeach-VLA-3B & \textbf{0.12} & \textbf{0.27} & \underline{0.51} & \textbf{0.30} \\
\bottomrule
\end{tabular}~\label{tab:nuscenes_impro}
\end{table*}

\clearpage
\section{System Prompt}
\label{sec:SystemPrompt}

DriveTeach-VLA involves four key system prompts, each serving a distinct purpose in DVD pretraining, CoT pseudo-labeling, language instruction for TGP-Prompter, and language instruction for TGP-Planner.

\noindent \textbf{(1) Critical-object detection prompt.}
We first employ GroundingDINO to detect critical objects from the training set. These pseudo-labeled objects (Fig.~\ref{fig:sys_dino}) are used as the supervision signal for the teacher ViT in DVD stage. Here, the bounding box colors vary across categories. We follow the default GroundingDINO configuration to assign a distinct color to each class.

\noindent \textbf{(2) CoT pseudo-labeling prompt.}
To obtain reliable reasoning supervision, we generate large-scale pseudo CoT annotations using the Qwen2.5-VL-72B model. This is essential: without providing high-quality reasoning traces, RL tends to collapse. The CoT annotation process covers three tasks---\emph{critical-object auditing}, \emph{natural-language causal explanation}, and \emph{meta-behavior selection}. The system prompt used for generating these annotations is shown in Fig.~\ref{fig:sys_cot_label}.

\noindent \textbf{(3) Language instruction for TGP-Prompter.}
During TGP-Prompter SFT learning, the model is required to predict 2D-TGP trajectory. The prompt is in Fig.~\ref{fig:sys_sft_tgp} defining the format of 2D-TGP.

\noindent \textbf{(4) Language instruction for TGP-Planner.}  This prompt incorporates several structured CoT subtasks to instruct TGP-Planner to reason step-by-step under the guidance of 2D-TGP. The Prompt is in Fig.~\ref{fig:sys_cot_rft}.

\begin{figure*}[ht]
\begin{tcolorbox}[enhanced jigsaw,width=\textwidth]
\textbf{Critical Object Detection via Grounding Dino}

car. truck. bus. trailer. construction\_vehicle. pedestrian. motorcycle. bicycle. barrier. traffic\_cone. traffic\_light. traffic\_sign.
\end{tcolorbox}
\caption{The critical object to detect using Grounding Dino.}
\label{fig:sys_dino}
\end{figure*}

\begin{figure*}[ht]
    
\begin{tcolorbox}[enhanced jigsaw,width=\textwidth]
\tiny{
You are an expert driver. Suppose you are driving. Let's complete the following tasks step by step.

\textbf{Input} 

- 1 frame of front-view image collected from the ego-vehicle at the present timestep. Picture 1: <image> the front view of the ego-vehicle

 - Current high-level intent (string): \{GO STRAIGHT, TURN RIGHT, TURN LEFT\}

- 1.5-second past trajectory (3 steps at 2 Hz), each trajectory point format (x:float, y:float, heading:float):    - t-3: (x1, y1, h1),   
- t-2: (x2, y2, h2),    - t-1: (x3, y3, h3).

- 4-second expert trajectory (8 steps at 2 Hz), each trajectory point format (x:float, y:float, heading:float):    - t+1: (x1, y1, h1),   - t+2: (x2, y2, h2),   - t+3: (x3, y3, h3),  - t+4: (x4, y4, h4),  - t+5: (x5, y5, h5),  - t+6: (x6, y6, h6),   - t+7: (x7, y7, h7),  - t+8: (x8, y8, h8).

\textbf{Task 1: Critical Objects and Conditions Detection.} Decide whether at least one critical instance of each class could influence the ego-vehicle's future path (no omissions). A vehicle can be a car, bus, truck, motorcyclist, scooter, etc. traffic\_element includes traffic signs and traffic lights. road\_hazard may include hazardous road conditions, road debris, obstacles, etc. A conflicting\_vehicle is a vehicle that may potentially conflict with the ego’s future path. Output "yes" or "no" for every class (no omissions).    Object classes to audit:      - nearby\_vehicle      - conflicting\_pedestrian      - cyclist      - construction      - traffic\_element      - weather\_condition      - road\_hazard      - emergency\_vehicle      - animal      - special\_vehicle      - conflicting\_vehicle      - door\_opening\_vehicle 

\textbf{Task 2: Natural Language Explanation.}  Compose a concise natural-language description of the 4-second expert trajectory for the ego vehicle that the expert driver (you) plans and explain why the expert driver plans to execute this trajectory.   - Mention only the classes you marked "yes" in the previous task.   - Describe how each of those critical objects or conditions influences the optimal trajectory.   - Do not invent objects or conditions not present in the input.

\textbf{Task 3: Meta-Behaviour Selection.} Assign exactly one category from each list. Choose the label that best summarises the overall behaviour of the expert trajectory:   - speed  { keep, accelerate, decelerate }   - command  { straight, yield, left\_turn, right\_turn, lane\_follow, lane\_change\_left, lane\_change\_right, reverse }   Choose the label that best summarises the overall behaviour of the expert trajectory.   - If none fits, use ‘other‘, but do this sparingly. 

\textbf{Output format (strict JSON, no extra keys, no markdown codeblock chars(```), no commentary)}

\{   "critical\_objects":

~~~~    \{  "nearby\_vehicle": "yes | no",     

~~~~  "conflicting\_pedestrian": "yes | no",     

~~~~  "cyclist": "yes | no",    

~~~~  "construction": "yes | no",     

~~~~  "traffic\_element": "yes | no",     

~~~~  "weather\_condition": "yes | no",     

~~~~  "road\_hazard": "yes | no",     

~~~~  "emergency\_vehicle": "yes | no",

~~~~"animal": "yes | no",    

~~~~"special\_vehicle": "yes | no",     

~~~~"conflicting\_vehicle": "yes | no",

~~~~"door\_opening\_vehicle": "yes | no"   \},

"explanation": "100-word description that references only the classes marked ’yes’", 

"meta\_behaviour": \{   

~~~~"speed"$\in$ \{keep, accelerate, decelerate, other \}, 

~~~~"command"$\in$ \{straight, yield, left\_turn, right\_turn, lane\_follow, lane\_change\_left, lane\_change\_right, reverse, other\}

\},   

\} 
}
\end{tcolorbox}
\caption{\textbf{Prompt for Qwen2.5-VL-72B to pseudo-label CoT reasoning stesps.}}
\label{fig:sys_cot_label}
\end{figure*}

\begin{figure*}[ht]
\begin{tcolorbox}[enhanced jigsaw,width=\textwidth]
\tiny{
You are an expert driver. Suppose you are driving. Given the input, predict 8 future 2D image trajectory, where x (key point x-coordinate): int, y (key point y-coordinate): int, h (heading): float, and 2D image trajectory format is [\{"point\_2d": [x1, y1, h1]\}, \{"point\_2d": [x2, y2, h2]\}, \{"point\_2d": [x3, y3, h3]\}, \{"point\_2d": [x4, y4, h4]\}, \{"point\_2d": [x5, y5, h5]\}, \{"point\_2d": [x6, y6, h6]\}, \{"point\_2d": [x7, y7, h7]\}, \{"point\_2d": [x8, y8, h8]\}].  

\textbf{Input} 

- 1 frame of front-view image collected from the ego-vehicle at the present timestep. Picture 1: <image> the front view of the ego-vehicle

 - Current high-level intent (string) $\in$ \{GO STRAIGHT, TURN RIGHT, TURN LEFT\}

- 1.5-second past trajectory(3 steps at 2 Hz), each trajectory point format (x:float, y:float, heading:float):    - t-3: (x1, y1, h1),   
- t-2: (x2, y2, h2),    - t-1: (x3, y3, h3).

\textbf{Output Format} 

strict JSON, no extra keys, no markdown codeblock chars, no commentary: \{   "future\_trajectory": "<answer>\{"point\_2d": [x1, y1, h1]\}, \{"point\_2d": [x2, y2, h2]\}, \{"point\_2d": [x3, y3, h3]\}, \{"point\_2d": [x4, y4, h4]\}, \{"point\_2d": [x5, y5, h5]\}, \{"point\_2d": [x6, y6, h6]\} \{"point\_2d": [x7, y7, h7]\}, \{"point\_2d": [x8, y8, h8]\}</answer>"\}
}
\end{tcolorbox}
\caption{Language instruction for TGP-Prompter to regress 2D-TGP trajectory.}
\label{fig:sys_sft_tgp}
\end{figure*}
\begin{figure*}[t]

\begin{tcolorbox}[enhanced jigsaw,width=\textwidth]
\tiny{You are an expert driver. Suppose you are driving. Let's complete the following tasks step by step.

\textbf{Input} 

- 1 frame of front-view image collected from the ego-vehicle at the present timestep Picture 1: <image> the front view of the ego-vehicle

 - Current high-level intent (string): \{GO STRAIGHT, TURN RIGHT, TURN LEFT\}

- 1.5-second past trajectory(3 steps at 2 Hz), each trajectory point format (x:float, y:float, heading:float):    - t-3: (x1, y1, h1),   
- t-2: (x2, y2, h2),    - t-1: (x3, y3, h3).

- To ensure a safe trajectory, you should pay attention to objects close to the key points, where x (key point x-coordinate): int, y (key point y-coordinate): int, and 8 key points \{"point\_2d": [x1, y1]\}, \{"point\_2d": [x2, y2]\}, \{"point\_2d": [x3, y3]\}, \{"point\_2d": [x4, y4]\}, \{"point\_2d": [x5, y5]\}, \{"point\_2d": [x6, y6]\}, \{"point\_2d": [x7, y7]\}, \{"point\_2d": [x8, y8]\}.

\textbf{Task 1: Critical Objects and Conditions Detection.  }  Decide whether at least one critical instance of each class could influence the ego-vehicle's future path (no omissions). A vehicle can be a car, bus, truck, motorcyclist, scooter, etc. traffic\_element includes traffic signs and traffic lights. road\_hazard may include hazardous road conditions, road debris, obstacles, etc. A conflicting\_vehicle is a vehicle that may potentially conflict with the ego’s future path. Output "yes" or "no" for every class (no omissions).    Object classes to audit:      - nearby\_vehicle      - conflicting\_pedestrian      - cyclist      - construction      - traffic\_element      - weather\_condition      - road\_hazard      - emergency\_vehicle      - animal      - special\_vehicle      - conflicting\_vehicle      - door\_opening\_vehicle 

\textbf{Task 2: Natural Language Explanation.}  Compose a concise natural-language description of the expert 4-second trajectory for the ego vehicle that the expert driver (you) plans and explain why the expert driver plans to execute this trajectory.   - Mention only the classes you marked "yes" in the previous task.   - Describe how each of those critical objects or conditions influences the optimal trajectory.   - Do not invent objects or conditions not present in the input.

\textbf{Task 3: Meta-Behaviour Selection.}  Assign exactly one category from each list. Choose the label that best summarises the overall behaviour of the expert trajectory:   - speed  { keep, accelerate, decelerate }   - command  { straight, yield, left\_turn, right\_turn, lane\_follow, lane\_change\_left, lane\_change\_right, reverse }   Choose the label that best summarises the overall behaviour of the expert trajectory.   - If none fits, use ‘other‘, but do this sparingly. 

\textbf{Task 4: Future Trajectory Prediction.}  Answer output should be wrapped in <answer>...</answer>
Given the input, critical objects/conditions, natural language explanation, meta-behaviour, and 2D-TGP key points (see Input), predict the optimal 4-second future trajectory (8 steps at 2 Hz) of the ego vehicle. Each trajectory point format (x:float, y:float, heading:float):    - t+1: (x1, y1, h1),   - t+2: (x2, y2, h2),   - t+3: (x3, y3, h3),  - t+4: (x4, y4, h4),  - t+5: (x5, y5, h5),  - t+6: (x6, y6, h6),   - t+7: (x7, y7, h7),  - t+8: (x8, y8, h8).

\textbf{Output format (strict JSON, no extra keys, no markdown codeblock chars(```), no commentary)}

\{   "critical\_objects":

~~~~    \{  "nearby\_vehicle": "yes | no",     

~~~~  "conflicting\_pedestrian": "yes | no",     

~~~~  "cyclist": "yes | no",    

~~~~  "construction": "yes | no",     

~~~~  "traffic\_element": "yes | no",     

~~~~  "weather\_condition": "yes | no",     

~~~~  "road\_hazard": "yes | no",     

~~~~  "emergency\_vehicle": "yes | no",

~~~~"animal": "yes | no",    

~~~~"special\_vehicle": "yes | no",     

~~~~"conflicting\_vehicle": "yes | no",

~~~~"door\_opening\_vehicle": "yes | no"   \},

"explanation": "100-word description that references only the classes marked ’yes’", 

"meta\_behaviour": \{   

~~~~"speed"$\in$ \{keep, accelerate, decelerate, other \}, 

~~~~"command"$\in$ \{straight, yield, left\_turn, right\_turn, lane\_follow, lane\_change\_left, lane\_change\_right, reverse, other\}

\},

"future\_trajectory": "<answer>- t+1: (x1, y1, h1),   - t+2: (x2, y2, h2),   - t+3: (x3, y3, h3),  - t+4: (x4, y4, h4),  - t+5: (x5, y5, h5),  - t+6: (x6, y6, h6),   - t+7: (x7, y7, h7),  - t+8: (x8, y8, h8).</answer>" 

\} 
}

\end{tcolorbox}
\caption{Language instruction for TGP-Planner under the guidance of 2D-TGP prompt.}
\label{fig:sys_cot_rft}
\end{figure*}

\end{document}